\documentclass[lettersize,journal]{IEEEtran}
\usepackage{graphicx}
\usepackage{amsmath}
\usepackage{amssymb}
\usepackage{booktabs}
\usepackage{threeparttable}
\usepackage[capitalize]{cleveref}
\usepackage{makecell}
\usepackage{multirow}
\usepackage{pifont}
\usepackage{float}
\usepackage[caption=false]{subfig}
\usepackage{adjustbox}
\usepackage{flushend}
\usepackage{color}
\usepackage{wrapfig} 
\usepackage{bm}
\usepackage{colortbl}
\usepackage{url}
\usepackage[T1]{fontenc}
\usepackage[dvipsnames,table]{xcolor}
\definecolor{RowColor}{rgb}{0.95, 0.95, 1}

\usepackage{algorithm}
\usepackage{algorithmicx}
\usepackage{algpseudocode}

\usepackage{newfloat}
\usepackage{listings}
\DeclareCaptionStyle{ruled}{labelfont=normalfont,labelsep=colon,strut=off} 
\floatstyle{ruled}
\newfloat{listing}{tb}{lst}{}
\floatname{listing}{Listing}

\definecolor{annotation}{RGB}{0, 153, 0}
\definecolor{key_words}{RGB}{236, 0, 141}
\newcommand{\cmark}{\textcolor{green}{\ding{51}}}
\newcommand{\xmark}{\textcolor{red}{\ding{55}}}

\usepackage{array}

\begin{document}

\title{PointGL: A Simple Global-Local Framework for \\ Efficient Point Cloud Analysis}

\author{Jianan Li$^{\star\dagger}$, Jie Wang$^{\star}$, and Tingfa Xu$^{\dagger}$
\thanks{*Equal contribution. $^{\dagger}$ Correspondence to: Jianan Li and Tingfa Xu.}
\thanks{J. Li, J. Wang and T. Xu are with Beijing Institute of Technology, Beijing 100081, China
        { \{lijianan,ciom\_xtf1\}@bit.edu.cn}}%
\thanks{J. Li and  T. Xu are also with the Key Laboratory of Photoelectronic Imaging Technology and System, Ministry of Education of China, Beijing 100081, China.}%
\thanks{T. Xu is also with  Chongqing Innovation Center, Beijing Institute of Technology, Chongqing 401135, China.}
}

\markboth{Journal of \LaTeX\ Class Files,~Vol.~14, No.~8, August~2021}%
{Shell \MakeLowercase{\textit{et al.}}: A Sample Article Using IEEEtran.cls for IEEE Journals}


\maketitle

\begin{abstract}
Efficient analysis of point clouds holds paramount significance in real-world 3D applications. Currently, prevailing point-based models adhere to the PointNet++ methodology, which involves embedding and abstracting point features within a sequence of spatially overlapping local point sets, resulting in noticeable computational redundancy. Drawing inspiration from the streamlined paradigm of pixel embedding followed by regional pooling in Convolutional Neural Networks (CNNs), we introduce a novel, uncomplicated yet potent architecture known as PointGL, crafted to facilitate efficient point cloud analysis. PointGL employs a hierarchical process of feature acquisition through two recursive steps. First, the \textit{Global Point Embedding} leverages straightforward residual Multilayer Perceptrons (MLPs) to effectuate feature embedding for each individual point. Second, the novel \textit{Local Graph Pooling} technique characterizes point-to-point relationships and abstracts regional representations through succinct local graphs. The harmonious fusion of one-time point embedding and parameter-free graph pooling contributes to PointGL's defining attributes of minimized model complexity and heightened efficiency. Our PointGL attains state-of-the-art accuracy on the ScanObjectNN dataset while exhibiting a runtime that is more than $\bm{5}$ times faster and utilizing only approximately $\bm{4\%}$ of the FLOPs and $\bm{30\%}$ of the parameters compared to the recent PointMLP model. The code for PointGL is available at \url{https://github.com/Roywangj/PointGL}.

\end{abstract}


\section{Introduction}
Acquiring concise geometric features from point clouds stands as a pivotal stride across a spectrum of 3D tasks~\cite{deng2022superpoint,zhao2022divide,chen2022pq} and multimedia applications~\cite{yang2020predicting,qiu2021geometric,han2022dual,shabbir2021enhancing}. Amid the gamut of methods for point cloud analysis, point-based models have been exhaustively investigated owing to their adept equilibrium between precision and efficiency. Noteworthy among these is PointNet++~\cite{qi2017pointnet++}, which is hailed as the progenitor in this domain. Although subsequent endeavors have predominantly fixated on enhancing precision, these enhancements frequently exact a toll in terms of augmented computational intricacy.
Hence, our study strives to fashion a point-based model that amalgamates simplicity and potency, thereby facilitating expeditious analysis of point clouds.

Contemporary point-based models conventionally embrace the PointNet++ pipeline. This pipeline encompasses the grouping of input points into local sets and subsequently executes point-wise embedding procedures within these local point sets. The ensuing phase entails employing symmetrical aggregation functions to distill local geometric features within each distinct point set. Nonetheless, a comprehensive scrutiny of this pipeline has brought to the fore two intrinsic drawbacks that curtail computational efficiency.

The initial inadequacy we discerned pertains to the substantial overlapping of the resultant local point sets acquired through grouping within the 3D space. This overlap results in a singular input point being encompassed by multiple point sets concurrently. Given that the embedding of points is executed individually within each local point set, this scenario leads to the replication of point embeddings for the same point across diverse point sets. Although minor disparities may arise in these embeddings due to variances in the relative positions of input points, we posit that this process entails noteworthy redundant computations.

\begin{figure}[t]
    \centering
    \includegraphics[width=0.49\textwidth]{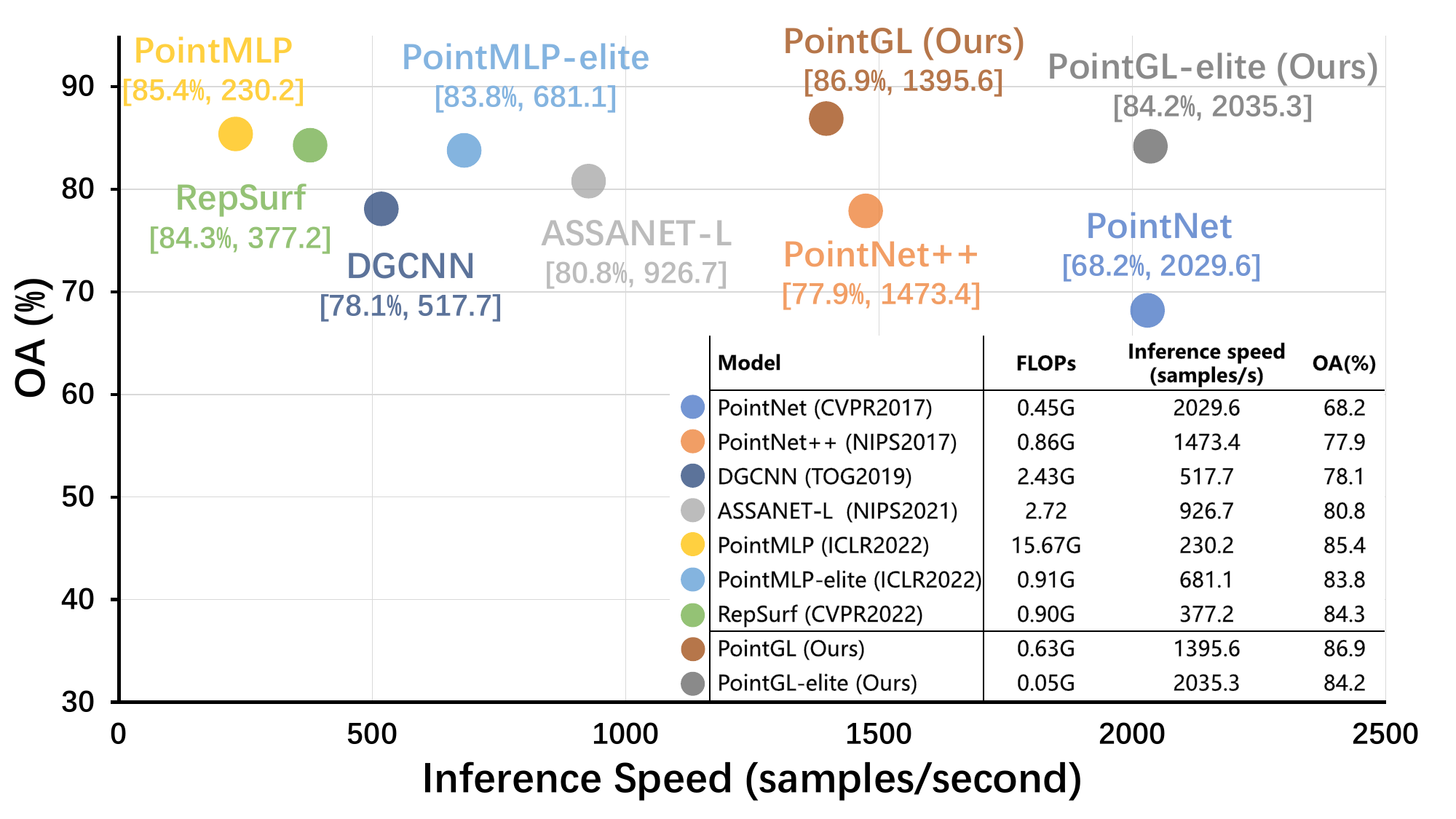}
    \caption{\textbf{Comparing Accuracy and Speed Among Various Approaches on the ScanObjectNN Dataset.} Our PointGL achieves the highest predictive accuracy and exhibits the fastest inference speed compared to alternative approaches.}
  \label{fig:acc-speed}
\end{figure}

The second inherent flaw is interconnected with the constraints posed by the process of extracting local features via point-wise embedding, succeeded by the application of symmetric functions like max pooling. This methodology can potentially lead to the erosion of intricate geometry due to its relatively feeble depiction of point-to-point relationships. Despite subsequent endeavors aimed at ameliorating this concern through the integration of more intricate encoders, such as convolutional schemes~\cite{wu2019pointconv,xu2021paconv}, graph-based techniques~\cite{wang2019dynamic}, or self-attention mechanisms~\cite{guo2021pct,zhao2021point}, these enhancements frequently entail a marked escalation in computational expenditure. As a result, it becomes imperative to identify more efficient approaches for the delineation and amalgamation of point-to-point relationships while upholding a commendable level of precision.

The abovementioned constraints impel us to explore a novel, efficient point-based paradigm guided by a concise design philosophy. Our design is grounded in a pivotal insight that individualized per-point feature embedding may not be indispensably correlated with optimal performance, and may inadvertently give rise to superfluous computational efforts. To surmount this, we execute feature embedding for each input point prior to the grouping of sets. This approach entails a singular embedding operation per point, engendering notable computational economies. Nonetheless, a pivotal quandary arises from the isolation of feature embedding for each point, consequently overlooking the interdependencies amongst points. These interrelationships, however, play a momentous role in capturing spatial geometric attributes.

In light of this, we introduce an additional step subsequent to the per-point embedding process, which intricately interlinks points within a designated local region and encapsulates their correlations within the local representation. To accomplish this task while preserving efficiency, we advocate for the incorporation of a graph structure, as it is adeptly suited for delineating interactions among nodes (points), thereby ensuring efficiency by streamlining the computation of edge features (relations).

Building upon the previously elucidated concepts, we have devised a novel architecture termed PointGL, tailored for the efficient analysis of point clouds. PointGL elucidates hierarchical features from the input point cloud via the integration of multiple progressive learning stages. Each of these stages encompasses two distinct phases: a \textit{Global Point Embedding} phase and a \textit{Local Graph Pooling} phase. The former assumes responsibility for the embedding of features per point, achieved by characterizing each point through elementary residual multi-layer perceptron (MLP) blocks. On the other hand, the latter encapsulates regional features by establishing connections amongst points within a designated local region, effectively constituting a rudimentary graph.
To ensure streamlined operations, we endow the graph edges with difference features derived from the disparities between the regional center point and its adjoining neighbors. Remarkably, our investigations have unveiled the remarkable informativeness of these edge features in delineating regional point-to-point relationships. Subsequently, these features are amalgamated via straightforward max pooling techniques, culminating in the extraction of regional representations.

Significantly, the process of local graph pooling is distinguished by its concise and efficient workflow, requiring minimal introduction of additional parameters. This inherent efficiency stems from the meticulous orchestration of global point embedding and local graph pooling, culminating in a streamlined and proficient framework for analysing point clouds. This framework not only excels in delivering high predictive accuracy but also exhibits notable swiftness in computational execution.

We have undertaken an extensive evaluation of PointGL across multiple benchmark datasets. As illustrated in \cref{fig:acc-speed}, our PointGL framework attains a state-of-the-art level of accuracy while concurrently delivering remarkable efficiency in inference speed. Noteworthy is the observation that our approach executes computations at nearly the same pace as PointNet++, yet remarkably enhances the overall accuracy by $\bm{9\%}$. Additionally, PointGL achieves accuracy levels comparable to those of PointMLP~\cite{ma2021rethinking}, yet operates more than $\bm{5}$ times faster, utilizing only about $\bm{4\%}$ of the Floating-Point Operations (FLOPs) and $\bm{30\%}$ of the parameters.

Moreover, we have extended our assessments to encompass segmentation and object detection, reaffirming the efficacy of PointGL in diverse downstream tasks. Furthermore, PointGL consistently showcases robustness in the face of point cloud corruptions. This resilience is substantiated by its attainment of state-of-the-art results across both the ModelNet-C and ShapeNet-C benchmarks, thereby solidifying its potential as a promising contender for real-world applications.

In summary, this study contributes in the following ways:
\begin{itemize}
    \item We introduce PointGL, which pioneers a novel and efficient point-based paradigm tailored for point cloud analysis. Despite its straightforward design, PointGL emerges as a potent tool for analyzing point clouds, thereby enriching the repertoire of existing point-based models.
    \item A novel local graph pooling operation is introduced, enabling the streamlined extraction of local geometric features. Notably, this operation demands minimal parameterization and seamless integration into pre-existing models.
    \item PointGL emerges as a lightweight model characterized by diminished computational intricacy. This attribute strikes an optimal equilibrium between precision and efficiency. Furthermore, its pronounced resilience in the face of corruptions reinforces its viability for real-world applications.
\end{itemize}

\section{Related Work}

\noindent\textbf{Learning on Point Clouds.}
The acquisition of discriminative features from point clouds constitutes a foundational endeavor for diverse 3D vision applications. Nonetheless, the inherent irregularity within point cloud data poses a challenge to conventional methodologies, such as voxel-based techniques~\cite{maturana2015voxnet,wu20153d}, and view-based approaches~\cite{guo2016multi,qi2016volumetric,su2015multi, saha2022translating, wiesmann2022retriever}. The former method involves projecting point clouds onto structured voxel grids, while the latter transforms them into 2D images. Both methodologies aim to capitalize on solutions developed for structured data, yet they suffer from a reduction of information due to the inherent projection process. In contrast, point-based techniques~\cite{qi2017pointnet,qi2017pointnet++,wang2019dynamic} tackle this issue by directly processing the raw point cloud data. PointNet~\cite{qi2017pointnet}, a pivotal advancement in 3D comprehension, employs multiple shared multi-layer perceptrons (MLPs) to learn individual point-wise features. This is achieved while maintaining permutation invariance through a max-pooling operator. Its successor, PointNet++~\cite{qi2017pointnet++}, enhances this approach by extracting both global and local geometric information through MLPs and a hierarchical architecture. This enables the extraction of more robust representations of point clouds. In this paper, we propose a more efficient and simplified hierarchical learning paradigm designed to capture the local geometric features of point clouds.

\noindent\textbf{Extraction of Local Geometric Features.}
The identification of local geometric features holds significant importance within the realm of point cloud analysis, a notion underscored by Liu et al. \cite{liu2020closer}. Previous investigations in this domain \cite{qi2017pointnet++, wang2019dynamic} have introduced diverse methodologies to address this challenge. For instance, PointNet++ \cite{qi2017pointnet++} has employed shared MLPs coupled with max pooling, while DGCNN \cite{wang2019dynamic} introduced EdgeConv to capture interpoint relationships and DC-GNN~\cite{meraz2022dcgnn} incorporates a channel dropout mechanism alongside a hierarchical feature selection strategy at each network layer for dynamic graph construction.
Alternative strategies \cite{thomas2019kpconv,zhao2021point} have explored deformable kernels, as exemplified by KPConv \cite{thomas2019kpconv}, or integrated attention mechanisms, as showcased in Point Transformer \cite{zhao2021point}, to model point interactions. Furthermore, PointMLP \cite{ma2021rethinking} utilizes a straightforward feed-forward residual MLP module accompanied by a geometric affine module to extract local features. This serves to demonstrate that a simplistic hierarchical MLP architecture can yield commendable performance while sidestepping the need for intricate local geometric extractors. In a similar vein, PointNext \cite{qian2022pointnext} employs an Inverted Residual MLP to abstract local features, yielding favorable results. Within this study, we propose a parsimonious yet efficacious \textit{local graph pooling} approach, which manifests exceptional prowess in abstracting local features. Notably, this comes at the expense of minimal supplementary parameters and computational overhead.

\noindent\textbf{Architectural Frameworks for Point Clouds.}
Within the realm of point cloud analysis, various architectural frameworks have emerged to process point cloud data. These include MLP-based approaches~\cite{qi2017pointnet,qi2017pointnet++,ma2021rethinking}, convolution-based methods~\cite{xu2018spidercnn,li2018pointcnn,wu2019pointconv}, graph-based models~\cite{wang2019dynamic,lin2021learning}, relation-based models~\cite{ran2021learning}, and transformer-based models~\cite{guo2021pct,zhao2021point}. While these architectures have demonstrated impressive accuracy, they often suffer from protracted inference times. In contrast, this paper advocates for a concise design philosophy, eschewing convoluted architectures in favor of presenting a streamlined yet potent framework for efficient point cloud analysis. This approach endeavors to strike a harmonious balance between accuracy and efficiency, a facet of particular significance in real-world applications.

\noindent\textbf{Robustness of Point Cloud Models.}
The assessment of the robustness of deep learning models is imperative prior to their deployment in real-world scenarios. In recent times, the research community has directed its efforts toward formulating benchmarks to gauge the resilience of models using 2D image datasets. Examples include ImageNet-C~\cite{hendrycks2019benchmarking}, Imagenet-V2~\cite{recht2019imagenet}, and ObjectNet~\cite{barbu2019objectnet}. Concerning 3D point cloud data, Ren et al.~\cite{ren2022benchmarking} have introduced a classification scheme for common 3D corruptions and subsequently conducted an extensive evaluation of extant point cloud classification models. Their findings underscore the susceptibility of prevailing point cloud models to corruptions. In contrast, the uncomplicated architectural framework advocated in this study exhibits commendable robustness in the presence of corruptions.

\begin{figure*}[t]
  \begin{center}
    \includegraphics[width=0.98\textwidth]{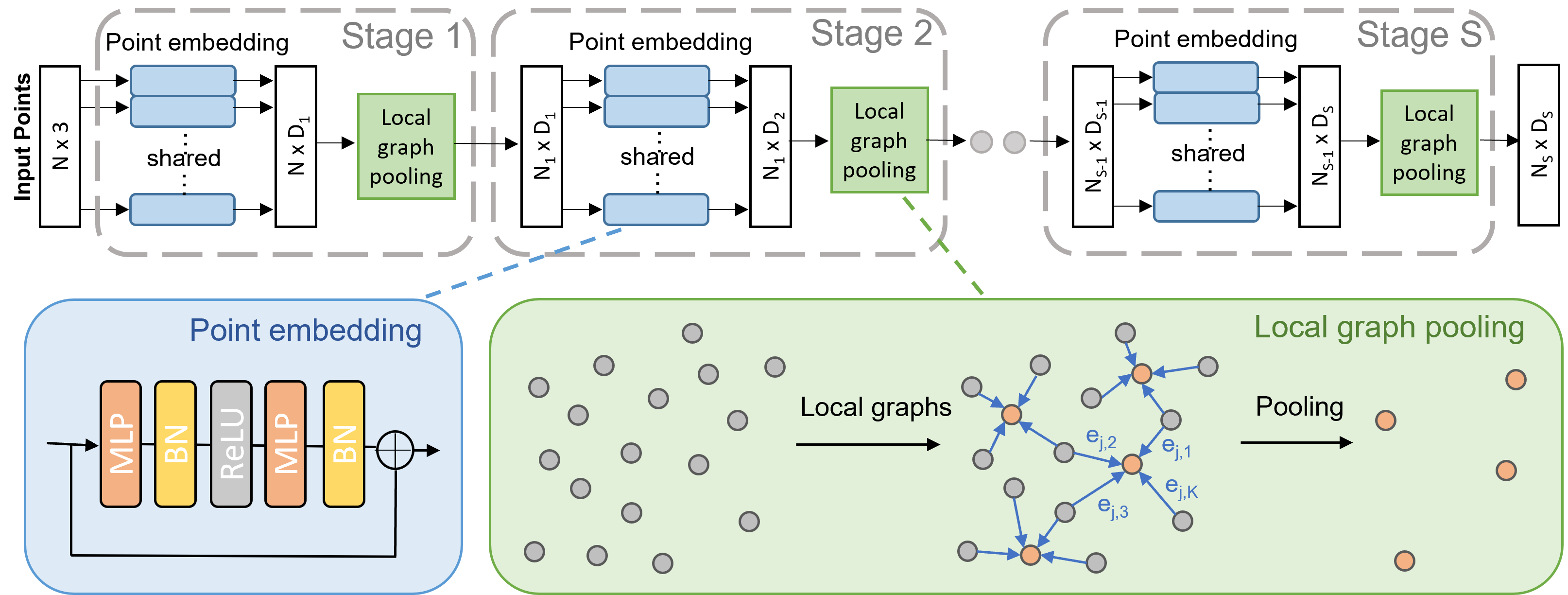}
  \end{center} 
    \caption{\textbf{Comprehensive Architecture of PointGL.} PointGL's architecture involves the extraction of hierarchical features from input point clouds through the stacking of multiple learning stages. Each stage initiates with a global point embedding phase, wherein feature embedding for individual points is conducted using residual MLP blocks. Following this, a local graph pooling phase captures and abstracts point-to-point relations into local representations by constructing a concise regional graph centered around each sampled point. The synergistic fusion of global point embedding and local graph pooling culminates in a coherent and efficient hierarchical framework for point cloud analysis.}
  \label{fig:overall}
  \vspace{-0.3cm}
\end{figure*}

\section{Method}

\subsection{Revisiting Point-based Approaches}
Point-based methods, notably exemplified by PointNet~\cite{qi2017pointnet} and PointNet++~\cite{qi2017pointnet++}, stand as pioneering instances in this domain, furnishing a foundational framework for subsequent endeavors. PointNet++ undertakes the hierarchical abstraction of intrinsic geometric features across multiple stages, facilitated by a collection of $\rm{N}$ points, each characterized by Cartesian coordinates denoted as $\bm{\mathcal{P}}= \left \{\bm{p}_i\in\mathbb{R}^{\rm{1}\times3}|i=1,\cdots, \rm{N}  \right \}$. Within the $s$-th stage, PointNet++ employs the farthest point sampling (FPS) algorithm to select ${\rm N}_s$ points. For each of these sampled points, $\rm{K}$ neighbors are queried, thereby constituting ${\rm N}_s$ local point sets. Subsequently, PointNet++ proceeds to acquire insights into the local patterns within each local point set through the expression:
\begin{equation}
    \bm{g}_i = \underset{j =1,\cdots, \rm{K}}{\bm{\mathcal{A}}}\left\{\bm{h}\left(\bm{f}_{i,j}\right)\right\},
\end{equation}
Here, $\bm{f}_{i,j}$ signifies the feature of the $j$-th neighboring point pertaining to the $i$-th sampled point and $\bm{g}_i$ denotes the aggregated local feature for the $i$-th sampled point. The function $\bm{h}(\cdot)$ is implemented by a multi-layer perceptron (MLP), thus conducting spatial encoding for a given point. Meanwhile, $\bm{\mathcal{A}}$ represents a symmetric function, like max pooling, employed to consolidate the encoded point features. Through the aggregation of multiple stages, featuring diminishing ${\rm N}_s$ and augmented spatial coverage within each local point set, PointNet++ progressively engenders abstracted local geometric features coupled with an expanding receptive field.

Subsequent point-based methodologies~\cite{xu2021paconv,liu2019relation,thomas2019kpconv,zhao2021point, wang2021papooling} have primarily directed their efforts towards enhancing the effectiveness of $\bm{h}(\cdot)$ to more adeptly capture interrelations between points, achieved by integrating convolution, graph, or attention mechanisms. Notably, the majority of these methodologies adhere to the overarching framework established by PointNet++: during each stage, input points are conglomerated into a sequence of local point sets, wherein feature embedding is executed for the points within each point set. Subsequently, local features are distilled through an aggregation function. However, a comprehensive scrutiny of this framework has illuminated two inherent limitations that could potentially impede computational efficiency.

Primarily, it is evident that during the $s$-th stage, there are ${\rm N}_{s-1}$ input points. Following the process of grouping, ${\rm N}_s$ local point sets emerge, with each assemblage accommodating $\rm{K}$ neighboring points. Typically, ${\rm N}_s$ constitutes merely half of ${\rm N}_{s-1}$, and $\rm{K}$ is selected from the range of $[16, 64]$, thus implying that the majority of input points are represented in multiple point sets.
Given that $\bm{h}(\cdot)$ operates on every point within each point set, on average, every input point necessitates a cumulative count of ${\rm N}_s  {\rm K} / {\rm N}_{s-1}$ feature embeddings. While point embeddings within distinct point sets might encompass varying spatial details, such as relative coordinates concerning a point set center, we posit that subjecting an individual point to multiple feature embeddings might not be pivotal for performance enhancement. However, this approach unavoidably introduces a substantial degree of computational superfluity. It's paramount to recognize that this impact becomes significantly magnified upon the integration of intricate spatial encoders in lieu of conventional MLPs.

Secondly, a noteworthy observation pertains to the fact that PointNet++ embarks on the abstraction of geometric features within a local point set via point-wise MLP, succeeded by a symmetric function. Regrettably, this approach is prone to diluting intricate geometric subtleties, primarily stemming from its limited capacity to model interpoint relationships. This shortcoming has prompted subsequent studies to address the concern through the introduction of intricate spatial encoders or aggregation functions. Yet, the unintended consequence of these endeavors has been the imposition of exorbitant computational demands and a significant expansion in memory usage, thus compromising overall efficiency.
Hence, the endeavor to enhance efficiency while upholding accuracy necessitates a more streamlined approach to the modeling and aggregation of interpoint relationships. These two limitations collectively impel us to embark on an exploration of a novel point-based paradigm.

\subsection{PointGL Framework}
The comprehensive architecture of our PointGL is elucidated in \cref{fig:overall}, detailing its process of learning hierarchical features from input point clouds through the stacking of a total of $\rm S$ learning stages. In the context of the $s$-th stage, the input is constituted by ${\rm N}_{s-1}$ points denoted as $\bm{\mathcal{P}}_{s-1}=\left \{(\bm{p}_{i}, \bm{f}_{i}) \ | \ i=1,\cdots,{\rm N}_{s-1}  \right \}$. Here, each point $i$ is characterized by its $xyz$ Cartesian coordinates represented as $\bm{p}_{i}\in \mathbb{R}^3$, along with an associated feature vector $\bm{f}_i \in \mathbb{R}^{{\rm D}_{s-1}}$. The outcome of this stage encompasses ${\rm N}_{s}$ sampled points denoted by $\bm{\mathcal{P}}_{s}=\left \{(\bm{p}_{i}, \bm{g}_{i}) \ | \ i=1,\cdots, {\rm N}_{s}  \right \}$. Within this set, each sampled point $i$ is distinguished by a feature vector $\bm{g}_i \in \mathbb{R}^{{\rm D}_{s}}$ that encapsulates the localized pattern surrounding it.

Our PointGL introduces a novel and efficient approach to learn local patterns through a dual-step process. The first step, known as \textit{Global Point Embedding}, entails the execution of feature embedding for each point within the input set $\bm{\mathcal{P}}_{s-1}$. Diverging from the methodology of PointNet++, which generates spatially overlapping local point sets and rigorously encodes features for each point within each point set, PointGL undertakes feature embedding solely once for every input point. This strategic divergence substantially mitigates redundant computations. Subsequently, in the subsequent step known as \textit{Local Graph Pooling}, points are selectively sampled from $\bm{\mathcal{P}}_{s-1}$, and the relationships between each sampled point and its neighboring points are meticulously modeled and collectively integrated to formulate localized representations. Remarkably, this entire sequence is accomplished exclusively through straightforward operations, effectively upholding computational efficiency. In the following sections, we proceed to provide an in-depth elaboration of both of these fundamental steps.

\noindent\textbf{Global Point Embedding.}
This phase entails the embedding of each input point within $\bm{\mathcal{P}}_{s-1}$ by training a function $\bm{\Phi}: \mathbb{R}^{{\rm D}_{s-1}}\rightarrow \mathbb{R}^{{\rm D}_{s}}$ dedicated to feature embedding. The feature embedding process is as follows:
\begin{equation}
    \bm{v}_i = \bm{\Phi} \left(\bm{f}_{i}\right), i =1,\cdots, {\rm N}_{s-1},
\end{equation}
where $\bm{v}_i$ is the embedded feature for the $i$-th point. Following the approach introduced in~\cite{ma2021rethinking}, we adopt the perspective of $\bm{\Phi}(\cdot)$ as a sequence of residual point $\mathrm{MLP}$ blocks. Precisely, this function is learned through a sequence of uniform residual $\mathrm{MLP}$ blocks. The composition of $\mathrm{MLP}$ encompasses a fully connected (FC) layer, batch normalization (BN), and rectified linear unit (ReLU) activation, succeeded by another layer of FC and BN.

The employment of point-wise $\mathrm{MLPs}$ imparts invariance to point permutations upon the function $\bm{\Phi}(\cdot)$, which is capable of approximating a diverse array of continuous functions. Additionally, the point embedding procedure necessitates minimal operations, primarily reliant on meticulously optimized feed-forward $\mathrm{MLPs}$. Furthermore, given that point embedding is executed solely once for each input point, the efficiency of our PointGL remains unimpeded even with the integration of more intricate or deeper $\mathrm{MLPs}$.

\noindent\textbf{Local Graph Pooling.}
The objective of this phase is to derive local patterns through the modeling and aggregation of interpoint relationships within a confined area. Recognizing the efficacy of graphs in representing sets of objects (nodes) and their interconnections (edges), a logical approach to capture point-to-point relationships within a limited region is to establish a local graph by connecting internal points. Within this construct, each node corresponds to a point, and the edges symbolize interactions between pairs of points. Through the amalgamation of all edge features within the local graph, we can effectively unearth local patterns.

To achieve this, we undertake the process through three distinct steps of local point association followed by feature aggregation:
\textit{i)} We initiate the process by sampling $\rm{N}_s$ points from the input data at the current stage utilizing the farthest point sampling (FPS) technique. Following this, we employ the k-nearest neighbors (kNN) algorithm to gather $\rm{K}$ neighboring points corresponding to each of the sampled points.
\textit{ii)} Subsequently, each sampled point is connected to its respective neighbors, effectively constituting a series of local graphs. These graphs encode the point-to-point relations through the incorporation of edge features.
\textit{iii)} In the final step, we subject the edge features of each local graph to a symmetric aggregation function. This operation yields the local feature representation.

The pivotal stride involves generating edge features capable of effectively encoding relationships among points while upholding efficiency. To accomplish this, we embrace a rudimentary yet remarkably efficacious strategy, encompassing the computation of the difference feature between a sampled point and each of its neighbors as the edge feature:
\begin{equation}
    \bm{e}_{j,k} = \bm{\alpha}\odot\left( \bm{v}_{j,k}  - \bm{v}_j \right) + \bm{\beta}.
\end{equation}
In this context, $\bm{v}_{j} \in \mathbb{R}^{{\rm D}_s}$ and $\bm{v}_{j,k} \in \mathbb{R}^{{\rm D}_s}$ signify the features of the $j$-th sampled point and its $k$-th neighboring point, respectively. The parameters $\bm{\alpha}, \bm{\beta} \in \mathbb{R}^{{\rm D}_{s}}$ are subject to learning, with the initial value of $\bm{1}$ and $\bm{0}$, respectively, while $\odot$ signifies the Hadamard product.

The local output representation for the $j$-th sampled point, denoted as $\bm{g}_j$, is derived by aggregating the edge features linked with all the edges emanating from this point. This aggregation is defined as:
\begin{equation}
    \bm{g}_j = \bm{\mathcal{A}} \left( \bm{e}_{j,k} | \ k =1,\cdots, \rm K\right),
\end{equation}
where $\rm{K}$ is the number of neighboring points. We opt for the employment of the max pooling operation as the aggregation function $\bm{\mathcal{A}}\left(\cdot\right)$ due to its efficacy and simplicity.

Beyond its simplicity and compactness, our approach to local graph pooling offers several salient advantages: \textit{i)} it retains invariance to the ordering of neighbors, thereby ensuring resistance to variations in point permutations; \textit{ii)} it refrains from involving intricate operations, thus ensuring commendable computational efficiency; \textit{iii)} it is nearly parameter-free, facilitating its seamless integration in a plug-and-play manner.

\begin{algorithm}[t]
\centering
\renewcommand{\arraystretch}{0.5}
\caption{Pseudo code of PointGL in a Pytorch-like style.} 
\begin{lstlisting}[language={Python}]
`{\textbf{Input}}:`
  xyz - [N, 3], coordinate of the `input` point cloud, where N denotes the number of points
  points - [N, C], feature of the `input` point cloud, where C denotes the dimension of the feature
  ns - number of the points selected by Farthest Point Sampling (FPS) algorithm
  K - number of the neighboring points by k-nearest neighbor (kNN) algorithm
  `$\alpha$` - affine transformation parameter, default 1.0
  `$\beta$` - affine transformation parameter, default 0.0

`{\textbf{Output}}:`
  new_xyz - [ns, 3], coordinate of the output point cloud, where ns denotes the number of points
  new_points - [ns, C'], feature of the output point cloud, where C' denotes the dimension of the feature

`{\textbf{Function}}` local_graph_pooling(xyz, points, ns, K, `$\alpha$`, `$\beta$`):
    # Select points by FPS algorithm
    fps_idx = furthest_point_sample(xyz, ns)
    new_xyz = gather(xyz, fps_idx)      
    new_points = gather(points, fps_idx)     
    
    # Find neighboring points by k-NN algorithm
    grouped_xyz, grouped_points = knn(query_xyz=new_xyz, support_xyz=xyz, feat=points, k_number=K) 

    # Generate edge features
    grouped_points = grouped_points - new_points
    grouped_points = `$\alpha$` * grouped_points + `$\beta$`
    
    # Aggregate local features
    new_points = grouped_points.`max`[0]    
    
    return new_xyz, new_points

`\textbf{In Each Processing Stage:}`
    # Step1: Global Point Embedding
    points = residual_MLPs(points)
    
    # Step2: Local Graph Pooling
    new_xyz, new_points = local_graph_pooling(xyz, points, ns, K, `$\alpha$`, `$\beta$`)

\end{lstlisting}
\label{code:PointGL}
\end{algorithm}

\newcolumntype{I}{!{\vrule width 1pt}}
\begin{table*}[h]
	\setlength{\tabcolsep}{6.0pt}
	 \renewcommand{\arraystretch}{1.5}
      \caption{\textbf{Architectural Specifications.} MLPs Specification: output channel count of fully connected (FC) layers. Pooling Specification: ($\rm{N}_s$: number of sampled points, $\rm{K}$: number of neighboring points).}
		\label{table:network_arc}
		\centering
		\footnotesize
		\begin{tabular}{c I c|c|c|c I c|c|c|c}
			\toprule[0.8pt]
		    Model & \multicolumn{4}{cI}{\textbf{PointGL}} & \multicolumn{3}{c}{\textbf{PointGL-elite}} \\
			\toprule[0.8pt]
			Stage & S$_1$ & S$_2$ & S$_3$ & S$_4$ 
			& S$_1$ & S$_2$ & S$_3$ & S$_4$ \\ 
			\hline
			\multirow{3}{*}{MLPs} & [128] & [256]  & [512]  & [1024] 
			 & [64]  & [128]  & [256] & [256]   \\
			& [128, 128] & [256, 256]  & [512, 512]  & [1024, 1024] 
			 & [16, 64]  & [32, 128]  & [64, 256] & [64, 256]   \\
			&  [128, 128] & [256, 256]  & [512, 512]  & [1024, 1024]  
			 & ~  & ~  & ~  & ~   \\		 
			\hline
			Pooling  & (512, 24)  & (256, 24)  & (128, 24)  & (64, 24) 
			 & (512, 24)  & (256, 24)  & (128, 24) & (64, 24) \\
			\hline
			Output & 512$\times$128 & 256$\times$256 & 128$\times$512  & 64$\times$1024 
			& 512$\times$64 & 256$\times$128 & 128$\times$256  & 64$\times$256 \\
			\bottomrule[0.8pt]
		\end{tabular}
\end{table*}

\begin{table*}[t]
    \begin{center}
\renewcommand\arraystretch{1.15}
\footnotesize
    \caption{\textbf{Performance Comparison on ScanObjectNN and ModelNet40 Datasets.} The table presents metrics including overall accuracy (OA, \%); mean per-class accuracy (mAcc, \%); parameter count (\#Params); and FLOPs. Additionally, we assess the processing speed of all methods in samples per second on a single GeForce RTX 3090 GPU and two cores of an Intel Xeon Gold 5218R CPU@2.10GHz. ($\dagger$: Multi-scale inference as per~\cite{liu2019relation}.)}
    \vspace{-2mm}
    \begin{threeparttable}
    \label{tab:classification}
    \tabcolsep=0.36cm
    \begin{tabular}{l|l|cc|cc|cccc}
    \Xhline{3\arrayrulewidth}
    \multirow{2}{*}{Method}      & \multirow{2}{*}{Input}  & \multicolumn{2}{c|}{~~~\bf ScanObjectNN ~~~} & \multicolumn{2}{c|}{~~~\bf ModelNet40 ~~~} & \multirow{2}{*}{\#Params} & \multirow{2}{*}{FLOPs} & \multirow{2}{*}{\shortstack{Train \\ Speed}} & \multirow{2}{*}{\shortstack{Infer \\ Speed}}\\ \cline{3-6}
    & &  \textbf{OA}(\%) & mAcc(\%) & \textbf{OA}(\%) & mAcc(\%)  &   &   &  &  \\
    \hline
    PointNet~\cite{qi2017pointnet}   & 1k P          & 68.2  & 63.4  & 89.2  & 86.0 & 3.47M & 0.45G & \textbf{1104.3} & 2029.6  \\  
    DGCNN~\cite{wang2019dynamic}   & 1k P            & 78.1  & 73.6  & 92.9  & 90.2 & 1.82M & 2.43G & 275.4 & 517.7   \\ 
    KPConv~\cite{thomas2019kpconv} & 7k P     & - & - & 92.9 & - & 14.30M & -  & 152.8    & 309.9  \\ 
    ASSANet (L)~\cite{qian2021assanet} & 1k P     & 80.8 & 77.7 & 92.9 & - & - & 2.72G  & 312.8   & 926.7  \\ 
    PointASNL~\cite{yan2020pointasnl}  & 1k P$^*$  & -  & - & 93.2 & -   & 10.1M   & 1.80G  & - & -  \\  
    MVTN \cite{hamdi2021mvtn} & multi-view      & 82.8 & - & 93.8 & \textbf{92.0} & 4.24M & 1.78G  & -  & -   \\
    PAConv$^\dagger$~\cite{xu2021paconv} & 1k P     & - & - & 93.9 & - & 2.44M & 1.68G & - & - \\ 
    RPNet~\cite{ran2021learning}  & 1k P$^*$     & - & - & 94.1 & - & 2.70M & 3.90G  & -  & -   \\ 
    CurveNet~\cite{xiang2021walk} & 1k P     & - & - & 93.8 & - & 2.14M & 0.66G & 126.8 & 278.9 \\  
    \hline
    PointNet++~\cite{qi2017pointnet++}  & 1k P     & 77.9 & 75.4 & 90.7 & 88.4 & 1.48M & 0.86G & 693.6 & 1473.4 \\  
    PointMLP~\cite{ma2021rethinking}  & 1k P     & 85.4 & 83.9 &  94.1 & 91.3 & 13.24M & 15.67G & 88.4  & 230.2 \\  
    PointMLP$^\dagger$~\cite{ma2021rethinking}  & 1k P     & 86.5 & 85.1 &  94.5 & 91.4 & 13.24M & 15.67G & 88.4  & 230.2\\  
    PointMLP-elite~\cite{ma2021rethinking}  & 1k P     & 83.8 & 81.8 & 93.6 & 90.9 & 0.72M & 0.91G & 324.9  & 681.1 \\  
    RepSurf-U~\cite{ran2022surface} & 1k P     & 84.3 & 81.3 & 94.4 & 91.4 & 1.48M & 0.90G  & 73.4  & 377.2   \\ 
    RepSurf-U$^\dagger$~\cite{ran2022surface} & 1k P     & 84.6 & 81.9 & \textbf{94.7} & \textbf{91.7} & 1.48M & 0.90G  & 73.4  & 377.2   \\ 
    \rowcolor{RowColor} PointGL  & 1k P   & \textbf{86.9}  & \textbf{85.2} & 93.0 & 90.4 & 4.16M & 0.63G & 600.7  & 1395.6 \\ 
    \rowcolor{RowColor} PointGL-elite & 1k P    & 84.2  & 82.2 & 92.6 & 89.8  & \textbf{0.49M} & \textbf{0.05G} & 1103.5  & \textbf{2035.3} \\ 
    \Xhline{3\arrayrulewidth}
    \end{tabular}
    \end{threeparttable}
    \end{center}
    \vspace{-0.1in}
\end{table*}

\subsection{Architectural Details}
As elucidated earlier, each learning stage within our methodology samples a subset of points from the input, endowing each sampled point with local geometric information. Through the accumulation of multiple stages, this design progressively furnishes a smaller number of points, each enriched with an extended contextual awareness.

The PointGL network is structured with $\rm S=4$ stages. These stages exhibit escalating embedding dimensions of ${\rm D}_{s} = \left \{128, 256, 512, 1024 \right \}$, complemented by a diminishing count of sampled points denoted as ${\rm N}{s} = \left \{ 512, 256, 128, 64 \right \}$. Each stage encompasses a residual MLP block, leveraging $\rm{K} = 24$ nearest neighbors for the purpose of local feature aggregation.

To further optimize efficiency, drawing inspiration from~\cite{ma2021rethinking}, we have introduced an advanced iteration of PointGL, denoted as PointGL-elite. This variant refines the feature embedding dimensions based on the original PointGL framework. For a comprehensive overview of the architectural nuances, please consult \cref{table:network_arc}. In addition, \cref{code:PointGL} shows the implementation of PointGL.

\section{Experiment}
\label{sec:experiments}
A comprehensive evaluation of the PointGL approach was undertaken across diverse benchmarks to gauge its effectiveness. Furthermore, we conducted experiments to evaluate the method's robustness in the face of corruptions, and performed ablation studies to validate both the chosen design principles and parameter configurations.

\subsection{Shape Classification on ScanObjectNN}

\noindent\textbf{Data and Setup.}
The fundamental assessment of point cloud analysis methods relies on 3D object classification. For our primary evaluation, we utilized the ScanObjectNN benchmark~\cite{uy2019revisiting}. This benchmark draws upon real-world object instances to provide an authentic assessment of the model's performance. The benchmark constitutes a multi-class classification task, which comprises a total of $2,902$ real-world point clouds from across $15$ distinct classes.

During the training phase, both the PointGL and PointGL-elite models underwent training for $250$ epochs utilizing the AdamW optimizer with a batch size of $32$.
The optimizer utilizes a learning rate of $0.002$ and weight decay of 0.05. 
The CosineAnnealingLR~\cite{loshchilov2016sgdr} scheduler is employed to decrease the learning rate to the minimum value of 1e-4, and the warm up epochs is set to 0. 
The evaluation metrics encompass overall accuracy (OA) and class-average accuracy (mAcc) calculated on the test set.

\begin{table*}[t]
\caption{Ablations on \textbf{global point embedding} and \textbf{local graph pooling} on the ScanObjectNN dataset.}
\label{tab:ablation}
\vspace{-0.1in}
    \subfloat[Perform point embedding before/after kNN. \label{tab:embed-position}]{
    \tabcolsep=0.34cm
    \footnotesize
    \begin{tabular}{l|cccccc}
        \toprule
        \multirow{2}{*}{Position} & \multirow{2}{*}{OA(\%)} & \multirow{2}{*}{mAcc(\%)} & \multirow{2}{*}{\#Params} & \multirow{2}{*}{FLOPs} & \multirow{2}{*}{\shortstack{Train \\ Speed}} & \multirow{2}{*}{\shortstack{Infer \\ Speed}} \\
        & & & & & & \\
        \midrule
        Pos-kNN & 86.2 & 84.0 & \textbf{4.16M} & 7.55G &  222.1 & 563.8 \\
        Pre-kNN (Ours) & \textbf{86.9} & \textbf{85.2} & \textbf{4.16M} & \textbf{0.63G} &  \textbf{600.7} & \textbf{1395.6}\\
        \bottomrule
    \end{tabular}} 
    \hspace{3mm}
    \subfloat[Number of embedding layers.
    \label{tab:depth}]{
    \tabcolsep=0.45cm
    \footnotesize
    \begin{tabular}{l|cc}
        \toprule
         Depth& OA(\%)  & mAcc(\%)   \\
         \midrule
         16 layers & \textbf{86.9}  & \textbf{85.2} \\
         24 layers & 86.6  & 84.6 \\
         32 layers & 86.4  & 84.5 \\
         \bottomrule
    \end{tabular}} \\
    \subfloat[Component ablation.
    \label{tab:lgp-component}]{
    \tabcolsep=0.22cm
    \footnotesize
        \begin{tabular}{cccll}
        \toprule
        Max Pool & Local Graph & $\bm{\alpha} \mbox{-} \bm{\beta}$  & \multicolumn{1}{|l}{OA(\%)} & mAcc(\%) \\
        \midrule
        \cmark&\xmark &\xmark &\multicolumn{1}{|l}{ 82.1}  &  80.3\\
        \cmark&\cmark &\xmark & \multicolumn{1}{|c}{86.1 \textcolor{green!40!gray}{\footnotesize $\uparrow$4.0}} & 84.5 \textcolor{green!40!gray}{\footnotesize $\uparrow$4.2}\\    
        \cmark&\cmark &\cmark & \multicolumn{1}{|c}{\textbf{86.9} \textcolor{green!40!gray}{\footnotesize $\uparrow$0.8}} & \textbf{85.2} \textcolor{green!40!gray}{\footnotesize $\uparrow$0.7}\\
        \bottomrule
        ~ & ~ & ~ \\ 
    \end{tabular}}  
    \hspace{1.6mm}
    \subfloat[\scriptsize Number of neighboring points.
    \label{tab:abl_K}]{
    \tabcolsep=0.25cm
    \footnotesize
    \begin{tabular}{lcc}
        \toprule
         {\rm K} & \multicolumn{1}{|c}{OA(\%)}  & mAcc(\%)   \\
         \midrule
         12  & \multicolumn{1}{|c}{85.8}  & 83.7 \\
         24 & \multicolumn{1}{|c}{\textbf{86.9}} & \textbf{85.2} \\
         36  & \multicolumn{1}{|c}{86.4}  & 85.0 \\
         \bottomrule 
            ~ & \multicolumn{1}{c}{~}  & ~ \\
    \end{tabular}} 
    \hspace{1.6mm}
    \subfloat[Plug-and-play capability.
    \label{tab:lgp-plug}]{
    \tabcolsep=0.22cm
    \footnotesize
        \begin{tabular}{l|ll}
        \toprule
        Model & OA(\%)  & mAcc(\%) \\
        \midrule
        PointNet++ &77.9 &75.4 \\
        PointNet++ (LGP) &\textbf{78.7} \textcolor{green!40!gray}{\footnotesize $\uparrow$0.8}  &\textbf{75.7} \textcolor{green!40!gray}{\footnotesize $\uparrow$0.3} \\
        \hline
        PointMLP &85.4 &83.9 \\
        PointMLP (LGP) &\textbf{85.9} \textcolor{green!40!gray}{\footnotesize $\uparrow$0.5} &\textbf{84.3} \textcolor{green!40!gray}{\footnotesize $\uparrow$0.4} \\
        \bottomrule
    \end{tabular}}
    \vspace{-4mm}
\end{table*}

\noindent\textbf{Main Results.}
The \cref{tab:classification} furnishes a comparative analysis between PointGL and state-of-the-art techniques. To ensure a comprehensive evaluation of the various methods, an assortment of metrics was considered, encompassing accuracy and efficiency measurements such as classification accuracy, model complexity (parameter count and FLOPs), and processing speed. Notably, our PointGL exhibited superior levels of accuracy and efficiency within the purview of this benchmark assessment.

In a specific context, PointGL showcased a cutting-edge overall accuracy of $86.9\%$, asserting its dominance over preceding methodologies such as PointMLP~\cite{ma2021rethinking} and RepSurf~\cite{ran2022surface}. Moreover, our self-contained PointGL demonstrated superiority over both PointMLP and RepSurf, even when employing a multi-scale inference strategy~\cite{liu2019relation}. Of significant note is our approach's substantial performance advantage over PointNet++ by a substantial margin ($86.9\%$ \textit{vs.} $77.9\%$), thereby underscoring the supremacy of our local graph pooling mechanism for the abstraction of local patterns.

\noindent\textbf{Computational Efficiency.} 
Regarding computational efficiency, our PointGL operates at a pace that is nearly comparable to PointNet++, widely acknowledged as the swiftest point-based model with a hierarchical framework. When contrasted with the recent PointMLP, our PointGL boasts a mere $\bm{30\%}$ of the parameter count ($4.16 \rm{M}$ \textit{vs.} $13.24 \rm M$) and a mere $\bm{4\%}$ of the FLOPs ($0.63 \rm G$ \textit{vs.} $15.67 \rm G$), all while delivering over $\bm{5} \times$ the speed in inference ($1,395$ \textit{vs.}~$230$ samples/s).

In contrast to the more contemporary RepSurf-U, our PointGL experiences a slight increase in FLOPs, yet manages to achieve $\bm{7.2} \times$ and $\bm{2.7} \times$ faster speeds during training and inference, respectively. These findings underscore that a reduction in model complexity doesn't necessarily guarantee heightened efficiency. Our straightforward approach of global point embedding followed by local graph pooling effectively curtails redundant computations across local point sets, thereby serving as the fundamental basis for efficiency enhancement.

To further optimize efficiency, we introduce an accelerated version of PointGL termed PointGL-elite. With a mere $0.49 \rm M$ parameters, PointGL-elite significantly reduces FLOPs to just $0.05 \rm G$, accounting for only $\bm{5\%}$ of the FLOPs found in its PointMLP-elite counterpart ($0.91 \rm G$). Despite maintaining a competitive accuracy level of $84.2\%$ OA, PointGL-elite achieves an exceptional inference speed of $2,035$ samples/s, surpassing the speeds of PointMLP-elite and PointNet++ by nearly $\bm{3} \times$ and $\bm{1.4} \times$, respectively. These outcomes firmly establish PointGL-elite as the point-based model with minimal model complexity and the swiftest inference rate.

\subsection{Shape Classification on ModelNet40}
\noindent\textbf{Data and Experimental Setup.}
Furthermore, we conducted evaluations using the ModelNet40 benchmark~\cite{modelnet40}, which is a multi-class classification task that encompasses a collection of $9,843$ training and $2,468$ test CAD models, distributed across $40$ distinct categories. 
The training process for both PointGL and PointGL-elite involved employing the Stochastic Gradient Descent (SGD) optimizer over $300$ epochs with a batch size of $32$. The optimizer is configured with a learning rate of $0.1$, momentum of $0.9$ and weight decay of 2e-4. The learning rate is decayed to the minimum value of $0.005$ using the CosineLRScheduler scheme.

\noindent\textbf{Main Findings and Results.}
The comprehensive comparison with leading state-of-the-art methodologies is presented in \cref{tab:classification}. In the absence of utilizing a voting mechanism, our PointGL and PointGL-elite attain impressive overall accuracies of $93.0\%$ and $92.6\%$, respectively, notably surpassing PointNet++ by a considerable margin of $2.3\%$ and $1.9\%$. Although exhibiting a slight decrease in accuracy compared to CurveNet, our approach showcases substantial reductions in model complexity while simultaneously achieving enhanced efficiency. Specifically, PointGL-elite stands out with just $\bm{23\%}$ of the parameters and $\bm{8\%}$ of the FLOPs, while exhibiting training and inference speeds that are accelerated by $\bm{7.7}$-fold and $\bm{6.3}$-fold, respectively.

\subsection{Ablation Studies}

\noindent\textbf{Global Point Embedding.}
To validate our key observation that conducting per-point feature embedding within each local point set, akin to the PointNet++ approach, is unnecessary for achieving optimal performance and would introduce significant computational redundancy, we devised a model variant. This variant involves relocating the point embedding step to the local graph pooling stage, positioned immediately after the k-nearest neighbors (kNN) operation.

\cref{tab:ablation}\textcolor{red}{a} confirms that PointGL outperforms the model variant, achieving higher accuracy while significantly reducing model complexity and improving efficiency. Specifically, while maintaining an equivalent parameter count, PointGL reduces FLOPs by an impressive factor of $\bm{11} \times$ and demonstrates approximately $\bm{2.7} \times$ faster training speed, coupled with an inference speed enhancement of approximately $\bm{2.5} \times$. These results effectively validate the critical importance of the global point embedding design within the PointGL framework.

To examine the effect of the number of residual MLP blocks on point embedding and its consequent impact on performance, we manipulated the quantity of these blocks within each learning stage, creating PointGL variants with $16$, $24$, and $32$ layers by adjusting the number of blocks to $1$, $2$, and $3$, respectively. As illustrated in \cref{tab:ablation}\textcolor{red}{b}, the addition of extra embedding layers does not consistently result in improved accuracy.

\begin{figure}[t]
    \centering
    \includegraphics[width=0.45\textwidth]{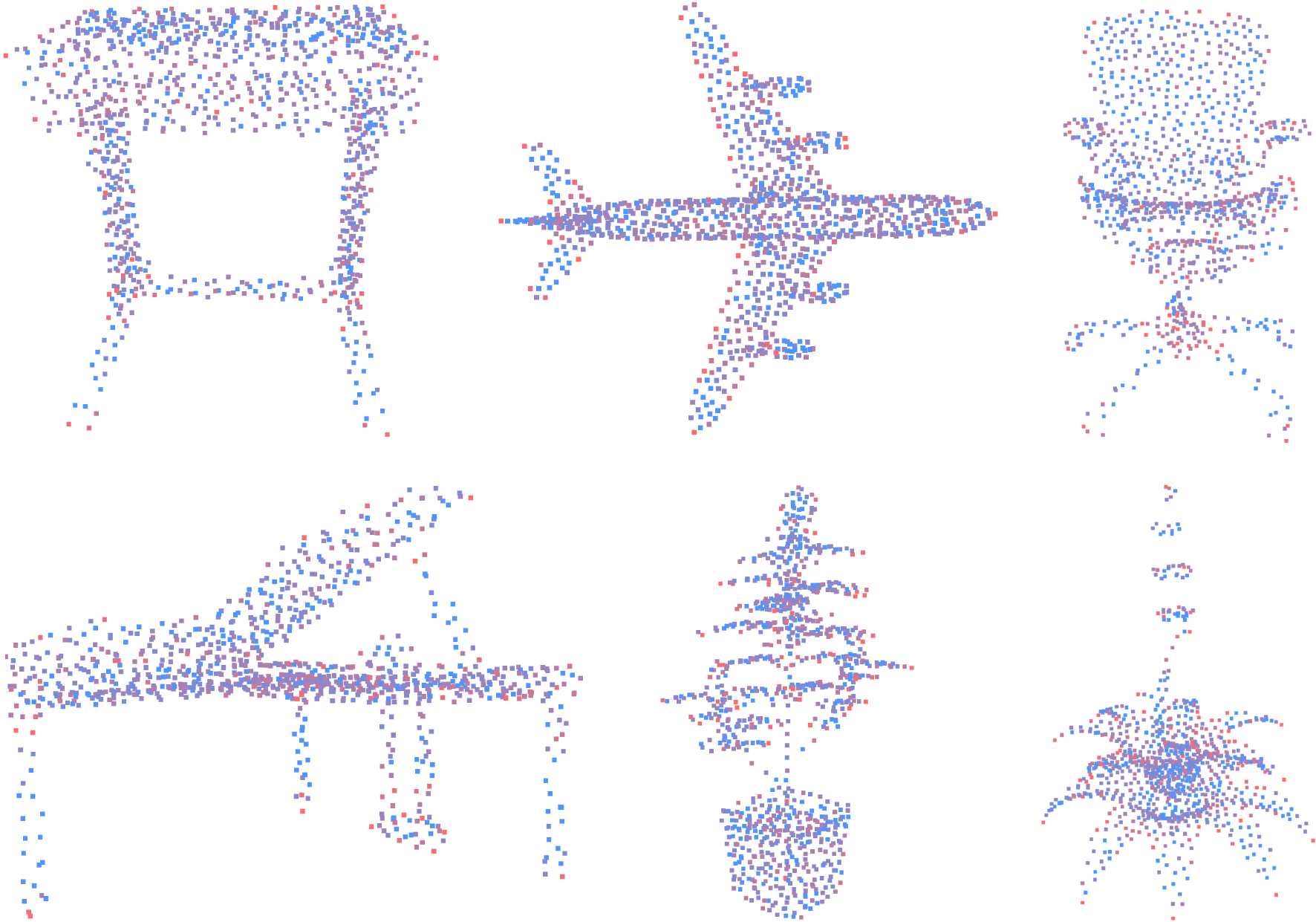}
    \caption{Salient geometric characteristics attained during the initial learning stage of PointGL on the ModelNet40 dataset. Each point's color corresponds to its received vote count, wherein a more intense \textcolor{red}{red} hue signifies a greater vote count, while a deeper \textcolor{blue}{blue} shade indicates a lower count of votes.}
   \label{fig:maxvis}
\end{figure}

\begin{table}[htbp]
\centering
\renewcommand\arraystretch{1.15}
\footnotesize
    \setlength{\tabcolsep}{13.0pt}
    \caption{Part segmentation results on the ShapeNetPart dataset.}
    \label{tab:part_segmentation}
    \begin{tabular}{l|cc|c}
        \toprule 
         Method& \makecell{Inst.\\mIoU(\%) }& \makecell{Cls.\\mIoU(\%) } & \makecell{Infer \\ Speed} \\
         \midrule
         PointNet~\cite{qi2017pointnet} &83.7 &80.4 &- \\
         DGCNN~\cite{wang2019dynamic} &85.2 &82.3 &-  \\
         KPConv~\cite{thomas2019kpconv} &86.4 &\textbf{85.1} &- \\
         GDANet~\cite{xu2021learning} & 86.5 & 85.0 & 100.8\\
         CurveNet~\cite{xiang2021walk} & \textbf{86.6} & - & 60.0\\
        \midrule
        PointNet++~\cite{qi2017pointnet++} &85.1 &81.9 &276.3 \\
        PointMLP~\cite{ma2021rethinking}  &86.1 &84.6  &179.6  \\
        \rowcolor{RowColor} PointGL    &85.6 &83.8  &284.6  \\
        \rowcolor{RowColor} PointGL-elite    &85.0 &83.0 &\textbf{309.0} \\
         \bottomrule
    \end{tabular}
\end{table}

\noindent\textbf{Local Graph Pooling.}
The findings from \cref{tab:ablation}\textcolor{red}{c} reveal the results of integrating each constituent aspect of local graph pooling into a foundational architecture that employs direct max pooling after the kNN operation. Evidently, the creation of graphs to encapsulate point-to-point relationships emerges as a crucial factor within PointGL, leading to a significant improvement of the base architecture's overall accuracy by $4.0 \%$. The inclusion of learnable parameters, specifically $\bm{\alpha}$ and $\bm{\beta}$, results in an additional enhancement of $0.8 \%$ in overall accuracy, resulting in a peak accuracy of $86.9 \%$. This strongly underscores the validity of our design decisions.

{Furthermore, we represent $\bm{\alpha} \in \mathbb{R}^{{\rm D}_s}$ and $\bm{\beta} \in \mathbb{R}^{{\rm D}_s}$ as ${\rm D}_s$-dimensional learnable vectors, employed for channel-wise adjustment of responses in differential features. To assess the efficacy of our design, ablative experiments were conducted by substituting both $\bm{\alpha}$ and $\bm{\beta}$ with learnable scalars. The experimental results indicate a noticeable performance degradation due to this modification, resulting in an overall accuracy reduction of $0.7\%$. These experiments affirm the effectiveness of the channel-wise feature adjustment design.}

{Next, we proceed to further validate the impact of the values of ${\rm K}$, representing the number of neighboring points in the kNN algorithm. As presented in \cref{tab:ablation}\textcolor{red}{d}, both excessively small (${\rm K}=12$) and large (${\rm K}=36$) values for ${\rm K}$ resulted in a decrease in predictive accuracy. This occurs because small ${\rm K}$ values may lead to a diminished receptive field for features, while large ${\rm K}$ values may attenuate local detailed features. Optimal predictive accuracy was achieved with a suitable value for ${\rm K}=24$. As a result, we adopted ${\rm K}=24$ in our experiments.}

We have successfully implemented the local graph pooling as a versatile drop-in block, distinguished by its minimal parameter requirement and enhanced efficiency. To assess its adaptability in point-based models like PointNet++ and PointMLP, we replaced the native max pooling with our local graph pooling, excluding the sampling and kNN operations. The results shown in \cref{tab:ablation}\textcolor{red}{e} demonstrate that the straightforward incorporation of local graph pooling consistently enhances performance across a range of models. This underscores the universal nature of our pooling approach, highlighting its seamless applicability within existing point-based models to improve feature aggregation, all while incurring minimal additional parameters and computational overhead.

\begin{figure}[t]
  \begin{center}
  \includegraphics[width=0.45\textwidth]{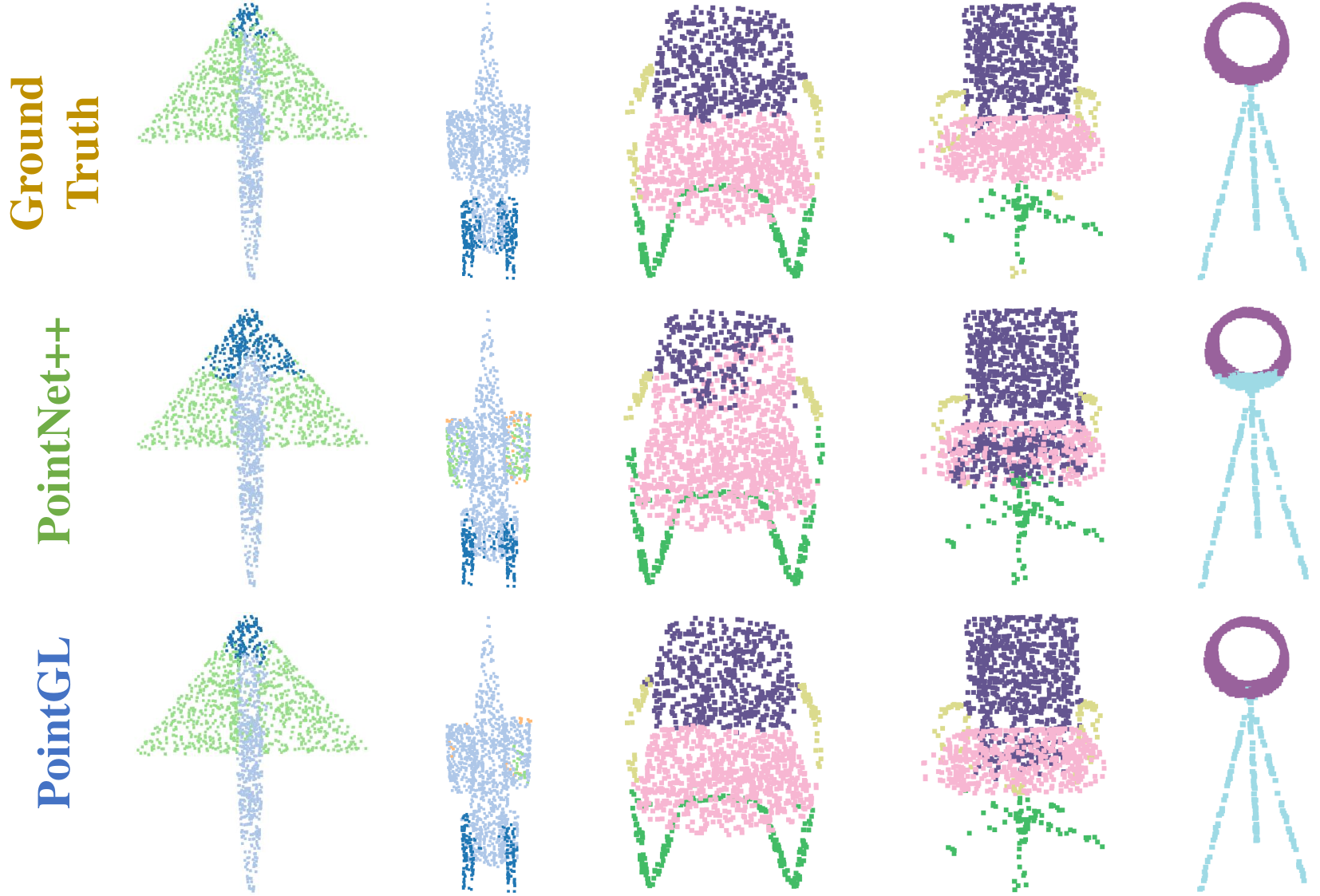}
  \end{center} 
  \caption{Visualization of part segmentation outcomes on the ShapeNetPart dataset. In contrast to PointNet++, PointGL's predictions exhibit a more robust alignment with the ground truth.}
  \label{fig:partseg}
\end{figure}

\begin{table}[t]
    \centering
    \renewcommand\arraystretch{1.15}
    \footnotesize
    \setlength{\tabcolsep}{4.5pt}
    \caption{Semantic segmentation results on the S3DIS Area-5 dataset.}
    \label{tab:semantic_segmentation}
    \begin{tabular}{l|ccc|ccc}
        \toprule 
         Method& \makecell{mIoU \\ (\%)} & \makecell{OA \\ (\%)} & \makecell{mAcc \\ (\%)} & \#Params & FLOPs & \makecell{Infer \\ Speed} \\
         \midrule
          PointNet~\cite{qi2017pointnet} & 41.1 & - & 49.0 &3.6M  &- &\textbf{205.6} \\ 
          DGCNN~\cite{wang2019dynamic} & 47.9 & 83.6 & -  &1.3M  &44.9G  &10.6 \\ 
         ASSANet-L~\cite{qian2021assanet} & 66.8 & - & - & 766.4M & 86.0G & 47.9 \\ 
         KPConv~\cite{thomas2019kpconv} & 67.1 & - & 72.8  &14.9M  &-  &- \\ 
         RepSurf-U~\cite{ran2022surface} & 68.9 & 90.2 & 76.0  &\textbf{1.0}M  &3.4G  &128.6 \\
         PT~\cite{zhao2021point} & \textbf{70.4} & \textbf{90.8} & \textbf{76.5} &7.8M  &3.1G  &47.2\\  
         \midrule
         PointNet++~\cite{qi2017pointnet++} & 53.5 & 83.0 & -  &3.0M  &6.5G  &169.4 \\ 
         \rowcolor{RowColor} PointGL    &65.6 &88.6 &71.9 &3.5M  &\textbf{2.5}G  &184.1 \\
    \bottomrule
    \end{tabular}
\end{table}

\begin{table*}[t]
\footnotesize
\setlength{\tabcolsep}{11pt}
\renewcommand\arraystretch{1.3}
    \caption{Performance on KITTI \textit{Validation} set. * denotes results from OpenPCDet.}
    \centering
    \begin{tabular}{l|lll|lll|lll}
        \toprule
        \multirow{1}{*}{\textbf{Method}} & \multicolumn{3}{c}{\textbf{Car (IoU=0.7)}} & \multicolumn{3}{c}{\textbf{Pedestrian (IoU=0.5)}} &  \multicolumn{3}{c}{\textbf{Cyclist (IoU=0.5)}} \\ 
        \cline{2-10}
        ~ & \textbf{Easy} & \textbf{Moderate} & \textbf{Hard} & \textbf{Easy} & \textbf{Moderate} & \textbf{Hard} & \textbf{Easy} & \textbf{Moderate} & \textbf{Hard} \\
        \midrule
        Second*~\cite{yan2018second} & 88.83  & 78.60  & 77.33  & 57.88  & 53.26  & 49.00  & 81.05  & 67.62  & 63.06  \\
        PointPillar*~\cite{lang2019pointpillars} & 87.05  & 77.29  & 75.65  & 56.49  & 50.83  & 46.84  & 79.56  & 63.35  & 59.53  \\
        Part-$A^2$-free*~\cite{shi2020points} & 89.07  & 78.64  & 78.10  & 68.02  & 63.13  & 58.32  & 86.70  & 72.33  & 69.59  \\
        \hline
        PointRCNN*~\cite{shi2019pointrcnn} & 88.57  & 78.53  & 77.75  & 60.38  & 53.44  & 49.36  & 87.89  & 73.29  & 67.65  \\
        \rowcolor{RowColor} + LGP & 88.72  & 78.55 \textcolor{green!40!gray}{$\uparrow$0.02} & 77.66  & 61.60  & 54.38 \textcolor{green!40!gray}{$\uparrow$0.94}  & 49.86  & 87.54  & 73.63 \textcolor{green!40!gray}{$\uparrow$0.34}  & 71.31 \\ 
        \hline
        PV-RCNN*~\cite{shi2020pv} & 89.58  & 83.19  & 78.86  & 63.72  & 57.35  & 53.40  & 84.93  & 71.67  & 68.29  \\
        \rowcolor{RowColor} + LGP & 89.43  & 83.49 \textcolor{green!40!gray}{$\uparrow$0.30}  & 78.86  & 65.87  & 59.35 \textcolor{green!40!gray}{$\uparrow$2.00}  & 54.56  & 86.36  & 72.67 \textcolor{green!40!gray}{$\uparrow$1.00}  & 69.09  \\
        \bottomrule
    \end{tabular}
    \label{tab:kittival}
\end{table*}

\begin{figure*}[t]
\centering
\includegraphics[width=1.0\textwidth]{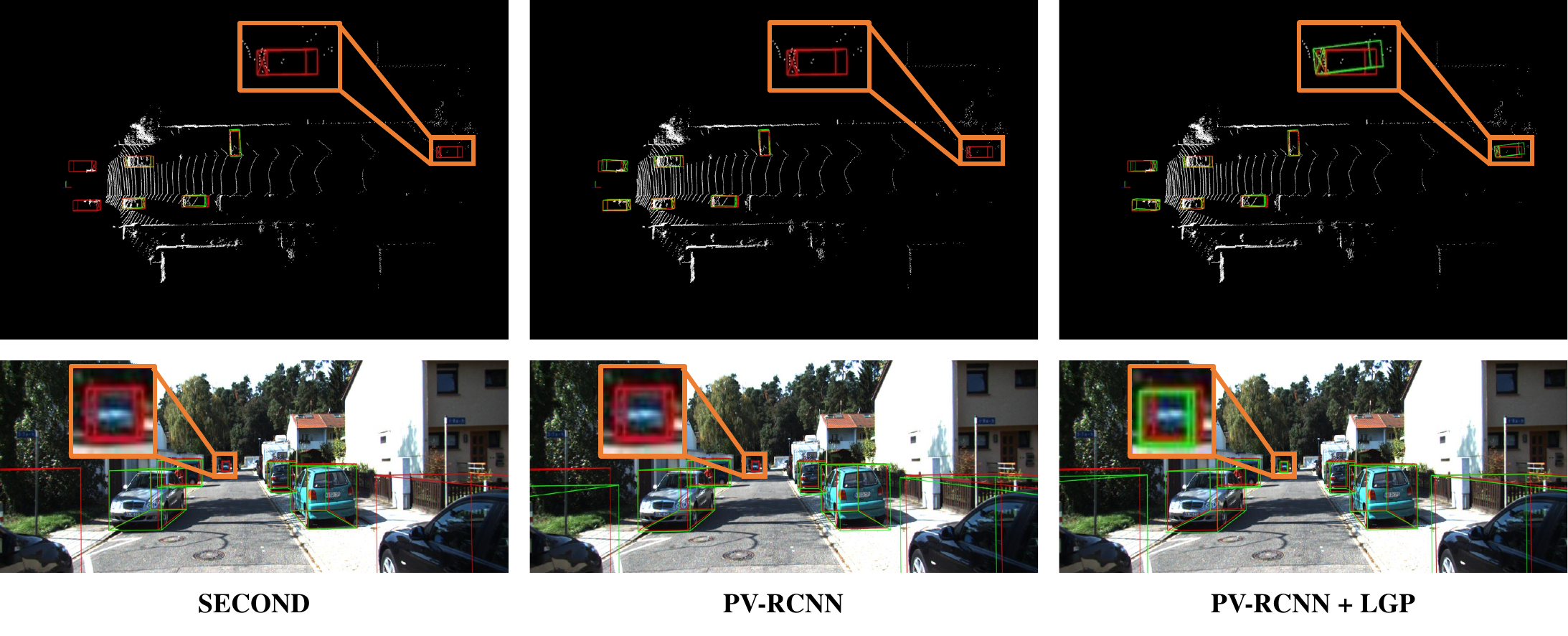}
\caption{Illustrative outcomes of 3D object detection on the KITTI dataset. Ground-truth and predicted objects are distinguished by \textcolor{red}{red} and \textcolor{green}{green} boxes, respectively. Integration of local graph pooling effectively recovers diminutive objects that were overlooked by the PV-RCNN baseline in intricate distant scenarios.}
\label{fig:detection_kitti}
\vspace{-3mm}
\end{figure*}

\noindent\textbf{Visualization of Features.}
To deepen our understanding of the behavior of our hierarchical network, \cref{fig:maxvis} offers a visual representation of the focal areas to which the network directs its attention during its initial learning stage. This was achieved by gathering the indices generated by the max pooling process within each local graph and subsequently consolidating these indices across all local graphs, resulting in votes for each input point. Points that contribute more significantly to the local representation, embodying recognized patterns, accumulate increased votes. The highlighted structural elements encompassing planes, lines, corners, and similar features affirm the network's proficiency in discerning and encapsulating crucial local geometric attributes.

\subsection{Object Part and Scene Segmentation}
The versatility of our PointGL framework allows for its extension to a variety of tasks. To assess its performance in the context of 3D shape part segmentation, we conducted experiments on the ShapeNetPart dataset~\cite{yi2016scalable}. This dataset comprises a collection of $16,881$ shapes distributed across $16$ distinct classes, each annotated with $50$ parts. We randomly sampled $2,048$ points as inputs~\cite{qi2017pointnet++} and trained the model for $350$ epochs using the Adam~\cite{kingma2014adam}  optimizer with a batch size of $32$.
The optimizer is configured with betas=(0.9, 0.999) and learning rate of $0.003$. The stepLR scheduler is also employed to decrease the learning rate with step size of $40$ and gamma of $0.5$.

As demonstrated in \cref{tab:part_segmentation}, our PointGL achieved remarkable performance, boasting an instance average mean Intersection over Union (mIoU) of $85.6\%$, while also maintaining notable inference speed. Furthermore, our approach outperformed the PointNet++ baseline in the majority of categories. \cref{fig:partseg} visually reinforces how PointGL's predictions closely align with the ground truth.

Moreover, our PointGL approach underwent evaluation on the S3DIS Area-5 dataset~\cite{armeni20163d}, which pertains to 3D semantic segmentation. 
In our experimental setup, we train the PointGL model with a batch size of 32 over 100 epochs, employing the AdamW optimizer. The optimizer is specifically configured with a learning rate of 0.01 and a weight decay rate of 1e-4. To manage the learning rate schedule, we utilize the CosineLRScheduler scheme, which reduces the learning rate progressively until it reaches a minimum value of 1e-5.
As presented in \cref{tab:semantic_segmentation}, PointGL achieved remarkably competitive performance, achieving an mIoU of $65.6\%$, surpassing established benchmarks like PointNet++ ($53.5\%$), while simultaneously performing on par with contemporary methods like RepSurf-U. Significantly, these achievements were reached with a lower FLOP count and maintained high-speed inference. These results underscore the adaptability of our PointGL approach and its potential for utilization across a spectrum of downstream tasks.

\subsection{Object Detection}
\noindent\textbf{Data and Setup.}
Our proposed \textit{Local Graph Pooling} operation seamlessly integrates into point cloud backbones, facilitating the extraction of intricate geometric features that can profoundly benefit downstream tasks, such as object detection. To demonstrate the effectiveness of our approach, we replaced the native set abstraction layer with our proposed module in prominent detection frameworks, specifically PV-RCNN~\cite{shi2020pv} and PointRCNN~\cite{shi2019pointrcnn}. We then evaluated the resulting object detectors on the widely used KITTI dataset~\cite{geiger2013vision}. The dataset comprises a training set of $7,481$ samples, conventionally split into $3,712$ and $3,769$ samples for training and validation, respectively. Our detector was trained for $80$ epochs using the Adam optimizer, initialized with a learning rate of $0.01$. The weight decay of the optimizer is set to $0.01$.
The evaluation metric employed was the per-class Average Precision (AP).

\noindent\textbf{Results.}
The outcomes of integrating our innovative \textit{Local Graph Pooling} operation into the PV-RCNN~\cite{shi2020pv} baseline are detailed in \cref{tab:kittival}. This integration resulted in significant improvements of $0.3$, $2.0$, and $1.0$ in AP for the Car, Pedestrian, and Cyclist categories, respectively. These results strongly emphasize the effectiveness of our approach in enhancing the acquired feature representation with intricate geometric details, which substantially contribute to the successful detection of objects of various scales. Moreover, the benefits of our proposed methodology extend beyond PV-RCNN, as it also yields improvements in other state-of-the-art detectors like PointRCNN, highlighting its versatile applicability across diverse detection frameworks.

\begin{table}[t]
    \centering
    \renewcommand\arraystretch{1.15}
    \footnotesize
    \setlength{\tabcolsep}{2.5pt}
    \caption{Classification performance under real-world corruptions on the ModelNet40-C dataset.}
    \label{tab:modelnet40-c}
    \begin{tabular}{l|c|c|ccccccc}
        \toprule 
         Method& OA(\%) & \textbf{mCE}$\downarrow$ &  Sca & Jit & D-G & D-L & A-G & A-L & Rot \\
         \midrule
        DGCNN~\cite{wang2019dynamic} & 92.6 & 1.00 & 1.00 & 1.00 & 1.00 & 1.00 & 1.00 & 1.00 & 1.00 \\
        PointNet~\cite{qi2017pointnet} &90.7 & 1.42 & 1.27 & \textbf{0.64} & 0.50 & 1.07 & 2.98 & 1.59 & 1.90\\
        PointNet++~\cite{qi2017pointnet++} &93.0 & 1.07 & 0.87 & 1.18 & 0.64 & 1.80 & 0.61 & 0.99 & 1.41\\
        RSCNN~\cite{liu2019relation} &92.3 & 1.13 & 1.07 & 1.17 & 0.81 & 1.52 & 0.71 & 1.15 & 1.48\\
        GDANet~\cite{xu2021learning} & 93.4 & 0.89 & 0.83 & 0.84 & 0.79 & 0.89 & 0.87 & 1.04 & 0.98 \\
        SimpleView~\cite{goyal2021revisiting} & \textbf{93.9} & 1.05 & 0.87 & 0.72 & 1.24 & 1.36 & 0.98 & 0.84 & 1.32\\
        PAConv~\cite{xu2021paconv} &93.6 & 1.10 & 0.90 & 1.47 & 1.00 & 1.01 & 1.09 & 1.30 & 0.97\\
        CurveNet~\cite{xiang2021walk} &{93.8} & 0.93 & 0.87 & 0.73 & 0.71 & 1.02 & 1.35 & 1.00 & \textbf{0.81}\\
        PCT~\cite{guo2021pct} &93.0 & 0.93 & 0.87 & 0.87 & 0.53 & 1.00 & 0.78 & 1.39 & 1.04\\
        RPC~\cite{ren2022benchmarking} &93.0 & 0.86 & \textbf{0.84} & 0.89 & \textbf{0.49} & \textbf{0.80} & 0.93 & 1.01 & 1.08\\
        \midrule
        \rowcolor{RowColor} PointGL    &93.4 & \textbf{0.77} & 1.01 & 1.15 & 0.60  & 1.19 & \textbf{0.24} & \textbf{0.26} & 0.94\\
         \bottomrule
    \end{tabular}
\end{table}

\noindent\textbf{Visualization.}
Qualitative results obtained through the application of our approach are vividly depicted in \cref{fig:detection_kitti}. This visualization distinctly illustrates instances where the PV-RCNN baseline faces challenges in detecting small objects, especially in scenarios with sparse point density. In contrast, the incorporation of our proposed \textit{Local Graph Pooling} operation leads to a significant improvement in detection performance in such scenarios. This empirical observation strongly emphasizes the effectiveness of our method in capturing crucial geometric details necessary for achieving accurate long-range object detection, even in situations characterized by limited dense point data.

\subsection{Robustness Analysis}
Real-world applications dealing with point cloud data often encounter challenges arising from sensor inaccuracies and complex scene structures, leading to data corruptions. Thus, the ability to handle point cloud corruptions effectively becomes crucial for practical applications. To assess the resilience and versatility of our proposed PointGL approach, we conducted a series of experiments on the ModelNet-C~\cite{ren2022benchmarking} and ShapeNet-C~\cite{ren2022pointcloud} datasets. Our evaluation covers tasks such as point cloud classification and part segmentation, conducted under various corruption scenarios.

\noindent\textbf{Results on ModelNet-C.}
We assess the robustness of our proposed PointGL approach using the ModelNet-C~\cite{ren2022benchmarking} point cloud classification corruption test suite. This suite encompasses seven distinct types of corruptions, each spanning five severity levels. This comprehensive assessment framework provides a rigorous evaluation of the model's inherent robustness. Our model is trained on the clean ModelNet40~\cite{modelnet40} dataset and subsequently evaluated on the ModelNet-C~\cite{ren2022benchmarking} test suite. Training utilizes the SGD optimizer with a batch size of $32$ and a learning rate of $0.1$, conducted over $300$ epochs.

\begin{table}[t]
\renewcommand\arraystretch{1.15}
\footnotesize
    \setlength{\tabcolsep}{0.7pt}
    \caption{Segmentation performance under real-world corruptions on the ShapeNet-C Dataset.}
    \label{tab:shapenet-c}
    \centering
    \begin{tabular}{l|c|cccccccccc}
        \toprule
        Method & mCE$\downarrow$ & Scale & Jitter & Drop-G & Drop-L & Add-G & Add-L & Rotate \\ 
        \midrule
        DGCNN~\cite{wang2019dynamic} & 1.00  & 1.00  & 1.00  & 1.00  & 1.00  & 1.00  & 1.00  & 1.00  \\
        PointNet~\cite{qi2017pointnet} & 1.18  & 1.08  & 1.05  & 0.98  & 1.13  & 1.39  & 1.17  & 1.44  \\
        PointNet++~\cite{qi2017pointnet++} & 1.11  & 0.95  & 1.08  & 0.86  & 1.98  & 0.89  & 1.08  & 0.95  \\
        PAConv~\cite{xu2021paconv} & 0.93  & 0.93  & 1.07  & 0.93  & 0.93  & 0.74  & 0.95  & 0.95  \\ 
        GDANet~\cite{xu2021learning} & 0.92  & 0.92  & 1.01  & 0.94  & 0.95  & 0.71  & 0.96  & 0.97  \\ 
        PT~\cite{zhao2021point} & 1.05  & 1.08  & 1.07  & 1.03  & 1.08  & 1.11  & 1.07  & 0.91  \\ 
        Point-MLP~\cite{ma2021rethinking} & 0.98  & 0.97  & 1.13  & 0.89  & 0.99  & 0.93  & 1.06  & 0.88  \\
        OcCo-DGCNN~\cite{wang2021unsupervised} & 0.98  & 0.96  & 1.07  & 0.96  & 1.02  & 0.94  & 1.00  & 0.89  \\
        Point-BERT~\cite{yu2022point} & 1.03  & 0.94  & 1.10  & 0.87  & 0.93  & 1.17  & 1.20  & 1.03  \\ 
        Point-MAE~\cite{pang2022masked} & 0.93  & 0.91  & 1.04  & 0.85  & 0.88  & 0.78  & 1.03  & 1.00  \\ 
        \midrule
        \rowcolor{RowColor} PointGL  & \textbf{0.82}  & 0.99  & 1.10  & 0.94  & 1.05  & 0.42  & 0.43  & 0.84 \\
        \bottomrule
    \end{tabular}
\end{table}

\cref{tab:modelnet40-c} presents a comprehensive summary of the evaluation results for point cloud classification models applied to the ModelNet-C dataset. The results highlight a significant observation - many state-of-the-art methods, known for their high Overall Accuracy (OA) on the clean ModelNet40 dataset, exhibit notable vulnerability when subjected to corruptions. In contrast, our proposed PointGL excels by achieving the lowest mean Corruption Error (mCE) of $0.77$ across all corruption scenarios. This accomplishment unquestionably demonstrates the robustness of our model. These findings underscore the substantial potential of PointGL in practical real-world scenarios.

\noindent\textbf{Results on ShapePart-C.}
The ShapeNet-C~\cite{ren2022pointcloud} benchmark is specifically designed to systematically assess the robustness of point cloud segmentation models across a range of corruptions. This benchmark encompasses seven distinct corruption categories, each calibrated at five different severity levels. To rigorously evaluate the robustness of our proposed PointGL, we conducted an extensive set of experiments on the ShapeNet-C~\cite{ren2022pointcloud} dataset, focusing on the task of part segmentation. 
Throughout the model training procedure, we employed the AdamW optimizer for a duration of $300$ epochs, with batch size of $16$. The optimizer's configuration comprised a learning rate of $0.0002$ and a weight decay coefficient of $0.05$. Furthermore, we applied the CosineLRScheduler scheme to modulate the learning rate, gradually reducing it to a minimum value of 1e-6. The warm-up epoch period is set to 10. 

The performance of our proposed PointGL on the ShapeNet-C benchmark is highlighted in \cref{tab:shapenet-c}. The achieved class-wise mIoU score of $0.82$ significantly outperforms the state-of-the-art PointMAE method by a substantial margin of $0.11$. Notably, PointMAE is a masked autoencoder (MAE) model trained on a substantial volume of unlabeled point cloud data. This noteworthy achievement underscores that PointGL effectively harnesses vital local information while preserving the integrity of semantic information. The demonstrated performance not only speaks to the robustness of our approach but also underscores its broad applicability across different tasks.

\section{Conclusion}
\label{sec:conclusion}
In this paper, we introduce PointGL, an architecture that combines simplicity and potency to employ a novel compact paradigm for efficient point cloud analysis. Our approach initiates by generating feature embeddings for individual points through residual MLPs. Subsequently, we introduce an innovative technique called local graph pooling, aimed at capturing regional features while minimizing extra learnable parameters and computational overhead. Our experiments on diverse benchmarks consistently demonstrate PointGL's superior performance compared to previous state-of-the-art models, achieving this with significantly reduced model complexity and heightened efficiency. We anticipate that our PointGL architecture will serve as an inspiration for the community to reevaluate efficient network design strategies tailored to point clouds.


\bibliographystyle{IEEEtran}
\bibliography{PointGL}

\vfill

\end{document}